\documentclass[runningheads]{llncs}
\usepackage{graphicx}
\begin{document}
\title{Question-Answering Model for Schizophrenia Symptoms and Their Impact on Daily Life using Mental Health Forums Data}
\titlerunning{Question-Answering Model on Data From Mental Health Forums}
%
\author{Christian Internò\inst{1,2}, Eloisa Ambrosini\inst{3}}
\authorrunning{Internó et al.}
%
\institute{University of Bielefeld, Bielefeld, Germany \and
\email{christian.interno@uni-bielefeld.de}\\ \and
\email{e.ambrosini2@campus.unimib.it }\\}
\maketitle              
\begin{abstract}
In recent years, there is strong emphasis on mining medical data using machine learning techniques. A common problem is to obtain a noiseless set of textual documents, with a relevant content for the research question, and developing a Question Answering (QA) model for a specific medical field. The purpose of this paper is to present a new methodology for building a medical dataset and obtain a QA model for analysis of symptoms and impact on daily life for a specific disease domain. The ``Mental Health'' forum was used, a forum dedicated to people suffering from schizophrenia and different mental disorders. Relevant posts of active users, who regularly participate, were extrapolated providing a new method of obtaining low-bias content and without privacy issues. Furthermore, it is shown how to pre-process the dataset to convert it into a QA dataset.
The Bidirectional Encoder Representations from Transformers (BERT), DistilBERT, RoBERTa, and BioBERT models were fine-tuned and evaluated via F1-Score, Exact Match, Precision and Recall. Accurate empirical experiments demonstrated the effectiveness of the proposed method for obtaining an accurate dataset for QA model implementation. By fine-tuning the BioBERT QA model, we achieved an F1 score of 0.885, showing a considerable improvement and outperforming the state-of-the-art model for mental disorders domain.

\end{abstract}
\section{Introduction}
In this historical period, more and more online platforms, websites, social networks, and forums are being used to share and discuss various topics. These can be a supportive tool where people with illnesses interact, presenting their daily problems, symptoms, and difficulties, by speaking through the forum, they feel free to express their thoughts without fear of judgement from the outside world. In fact, a simple search on the Internet is enough to realise the huge number of existing forums on the most diverse topics, which, especially in the medical field, relating to illnesses and discomforts, can lead to a huge number of first-hand testimonies. These data remain unused even though they have great potential and could also find application in the medical field. 

Many applications of Natural Language Processing (NLP) techniques for data mining can be applied. It can be observed that in the current literature \cite{c2,c4,c6,c9} the application of Question Answering Models (QA) is widely used.
QA models can answer questions given some context, and sometimes even without any context in an open-domain way. They can extract answer phrases from paragraphs, paraphrase the answer generatively, or choose one option out of a list of given options. It all depends on the dataset that was used for training, on the problem it was trained for and to some extent the neural network architecture \cite{c4}.

The QA models need to understand the structure of the language, have a semantic understanding of the context and the questions, have an ability to locate the position of an answer phrase, and much more. It is difficult to train models that perform these tasks.
Google's BERT \cite{c2} is a paradigm shift in natural language modelling, in particular because of the introduction of the pre-training and fine-tuning paradigms: after pre-training in an unsupervised way on a massive amount of text data, the model can be rapidly fine-tuned on a specific downstream task with relatively few labels. Domain adaptation for a QA model is possible  by creating a specific QA dataset, that contains many paragraphs of text, several questions related to the paragraphs, their answers, the index of the beginning of the answers in the paragraph and training the model on this domain-specific data \cite{c2}. 
Through BERT models pre-training on schizophrenia data, it is possible to develop a QA model that can be used to obtain specific answers to questions concerning the limitations, problems, symptoms and specific information of patients with schizophrenia who periodically participate in forums dedicated to them. 

The construction of a QA system in the medical field presents unique challenges: \textbf{i) Obtaining large datasets related to the domain of interest.} Obtaining medical data for analysis or extraction is not easy, given the difficulty of collecting this type of data, the specificity required for certain diseases, privacy regulations and the presence of bias in them. \textbf{ ii)Addressing the creation of an effective QA dataset.} Developing a QA dataset presents several complexities, such as the identification of topics of interest in the corpus, the annotation of questions and answers relevant to the domain of interest, with the possibility of having multiple answers. \textbf{ iii) Train a QA model in an efficient way.} To achieve good results with a QA model, it is crucial to obtain a good quality QA dataset. Generally, for this task it is relied on domain experts who manually annotate relevant questions and answers, but this process can be time-consuming and laborious. Failing to obtain a trained QA model with a high variance of relevant questions and answers. Inadequate pre-processing of the corpus could be problematic for data extraction via QA model.

The novelty of this research comes from three aspects:
\begin{itemize}
    \item[•] We present a new method for obtaining data on a specific medical disease to build a non-biased, easy to obtain, and without privacy issue dataset. Through web scraping on forums \footnote{\url{https://www.mentalhealthforum.net/}} dedicated to schizophrenia patients, is it possible to obtain relevant information regarding patients' daily problems and symptoms e.g. What they had to stop doing or can no longer do because of the disease, what most patients have in common when they suffer from hallucinations. Within the forums, people feel free to express their thoughts without the burden of judgment and avoiding problems and privacy restrictions when using this data. The fact that they do not have an interlocutor ready to judge them allows them to write without filters, which is very important and allows to have data that is as truthful as possible and not influenced by external agents. Forums are usually frequented by numerous people from different places, with different social backgrounds and experiences; consequently, the sample is very large and significant. Furthermore, forums charge a fee to participate and administer the appropriate use of them. This way it can be ensured that all extracted posts come from real people suffering from the disease.
    \item[•] We show how optimize and speed up the question-answering annotation process and how processing the corpus containing user posts with the application of the Latent Dirichlet allocation (LDA) for finding topics in the corpus. The questions and answers are annotated in the QA dataset based on the most relevant aspects contained in each topic resulting from the topics-analysis.
    \item[•] We show the results obtained with the application of BioBERT, RoBERTa, and DistillBERT models. We additionally show how to set up a pipeline for the use of a QA model, how whit the use of the Retriever \cite{c7}, a filter that can quickly examine the entire archive of documents and pass the most relevant documents to the response, a significant improvement in efficiency of the model can be achieved. We empirically demonstrate  how it is possible to boost the performance in the schizophrenia field of those model by fine-tuning with the  schizophrenia QA dataset, reaching state-of-the-art in the field of mental issues QA models.
\end{itemize}

The reminder of the paper is structured as follows:
in Section 2, recent relevant works are discussed, showing the problems related to the medical domain datasets and the data mining of medical data by QA models.
In Section 3.2 and 3.3, it is presented how the data of the participants of the schizophrenia forum was acquired and pre-processed, how the SQuAD dataset used was constructed and the pipeline structure of the QA system.
In Section 3.4, and 3.5 the methods used for the QA model are explained, how the domain adaptation was performed, and which metrics were used.
In Section 3.6 results and evaluations are presented, describing examples of the model's response to medical data from the forum.
In Section 4 conclusions regarding the research carried out, and possible future work are discussed.

\section{Related Work}
Relevant to our work is work offering methods for data mining in the medical field, in particular work demonstrating techniques for obtaining large, non-biased medical datasets, setting up QA datasets and implementing QA models.
\subsection{Data for the medical domain application}
As shown in \cite{c6} one option for obtain medical data is the use of clinical charts, but this approach has limitations in patient privacy or difficult access. It can also be seen how in \cite{c8} the use of recordings of doctor-patient dialogues was experimented with, again addressing the limitation of privacy and the difficulty in implementing such a method to obtain many data and in applying numerous pre-processing techniques. In \cite{c9} they use examination results as the source of data, but direct contact with patient statement is lost, implying a loss of relevant information. One of the most interesting methods is presented in \cite{c32}, which used QA forums to create a QA dataset, selected questions in the medical-biological field on ``Reddit'' and used the answers with the highest score.
The problem is the lack of statements directly from patients afflicted with the disease under consideration, and the lack of multiple answers for the same question. We decided to present a new method for constructing a QA dataset that had no noise in it, easy to obtain, and with data directly provided by the patients themselves by scraping conversations between schizophrenic patients from ``The Mental Health forum''.

\subsection{Development of a QA dataset}
As shown in \cite{c15,c16,c17,c18} the LDA method gives good results in identifying topics contained in text files. It was therefore decided to base the construction of our QA dataset specific to the schizophrenia domain on the identification of questions and answers to be annotated based on the results of the LDA method for Topics Analysis. Instead of to simply annotating questions of any type by volunteers directly on the unprocessed corpus, not divided into topics paragraphs and without a precise list of questions to annotate \cite{c32}. 
As shown in \cite{c33} a QA dataset construction was performed for the COVID-19 and trained the COVID-QA model, but it use of very long text documents and none n-way answers to singles questions. Unlike our method, that provides division based on the topics identified via LDA into small paragraphs containing forum users' posts. The same question is asked multiple times in the same paragraph since each post contains a different but always satisfactory answer.

\subsection{QA model for the medical field}
It is possible to observe in \cite{c11,c2,c13,c14} the use of pretrained BERT models to perform fine-tuning and create a QA model for a specific domain. Currently, there is a large number of pre-trained models for different specific domains, in \cite{c36,c37} perform fine-tuning using the BERT model for the schizophrenia domain. In this paper, we show different experiments with DistilBERT, RoBERTa, BioBERT, and the fine-tuned models RoBERTa and BioBERT with our QA dataset\cite{c39}. 
BioBERT model was selected as the benchmark regarding Question Answering task in the medical field. In \cite{c39} BioBERT is trained on a large dataset of medical records but is not trained specifically for a single disease such as schizophrenia.

\section{METHODOLOGY} 
In this section we present the methods used for the acquisition and pre-processing of schizophrenia data in order to be able to construct a schizophrenia QA dataset. The pipeline structure of the question-answer system and how it was used and optimised. We illustrate how the training of BERT models was conducted for the schizophrenia domain and the experiments conducted using three different models and fine-tuned version for the schizophrenia domain, including a model already pre-trained on bio-medical tasks: DistilBERT \cite{c29}, RoBERTa-base-squad2 \cite{c24} and BioBERT \cite{c39}. 

\subsection{Acquisition and pre-processing of medical data}

The original corpus consists of a list of 415602 posts with their respective IDs and dates posted by different users.
First, unnecessary words and characters were removed so that the aspect words were meaningful for the subsequent clustering. Stop words, punctuation and numbers were eliminated.
We have done a stematisation of words, making them lowercase, and a branching of words, resulting in a shortening at the root. For example, ``apple'' and ``apples'' both become ``appl'' and are treated as the same word in the vectorization step.

We decided to divide the corpus into paragraphs, grouping posts into different paragraphs according to the topic they belong to. We based these paragraphs on the topics obtained through the topics anlysis carried out via the Latent Dirichlet Allocation (LDA)\cite{c17}. LDA is an unsupervised probabilistic generative method for modeling a corpus, which identifies patterns in word frequency to probabilistically estimate the topics of documents and the words used in those topics\cite{c17}. It assumes each document is made up of several topics and similar topics use similar words.
Each document can be represented as a probabilistic distribution over latent topics and the distribution of topics in all documents shares a common Dirichlet prior. Each latent topic in the LDA model is also represented as a probabilistic distribution of words, and the word distributions of topics also share a common Dirichlet prior\cite{c40}. 

A total of 35 different topics were obtained and 300 different relevant aspects.
An illustration of the results of the analysis can be seen in Table 1, it is possible to observe examples of topics and related aspects obtained, and how they are correlated to problems and symptoms of patients with schizophrenia.
The forum posts were divided into paragraphs and grouped by topics, to facilitate the annotation of questions and answers for the development of the QA dataset. Furthermore, the division into paragraphs based on the topic results facilitates the work for the QA model to find answers in the corpus more efficiently.

\begin{table}
 \centering
\caption{Topics and Aspects.}\label{tab1}
\begin{tabular}{|l|l|}
\hline
 Topics &  Aspects \\
\hline
1 &  ``hallucinations'', ``afraid of'', ``memory'', ``problems'', ...\\
2 &  ``stop'', ``caffè'', ``nicotine'', ``drinking'', ... \\
3 &  ``pandemic'', ``suffering'', ``society'', ``family'', ...\\
4 &  ``food'', ``weight'', ``struggle'', ``being clean'', ...\\
\hline
\end{tabular}
\end{table}

\subsection{Development of a domain-specific QA dataset}
By examining the aspects for each of the 35 resulting topics through LDA, the most relevant aspects for each topic inside the respective paragraph were identified. To then select the questions and answers related to each Aspect. For example, in one of the topics, one of the most relevant aspects is ``Afraid of'', in this case the question ``What is a schizophrenic afraid of?'' and the related answers found in the paragraph representing a specific topic were annotated; If the aspect is ``drinking'', the question will be formulated according to the context of the post e.g. ``What does a schizophrenic stop with?''. The diagram of this method is shown in Figure 1.

\begin{figure*}[tb]
    \centering
    \includegraphics[width=0.75\linewidth]{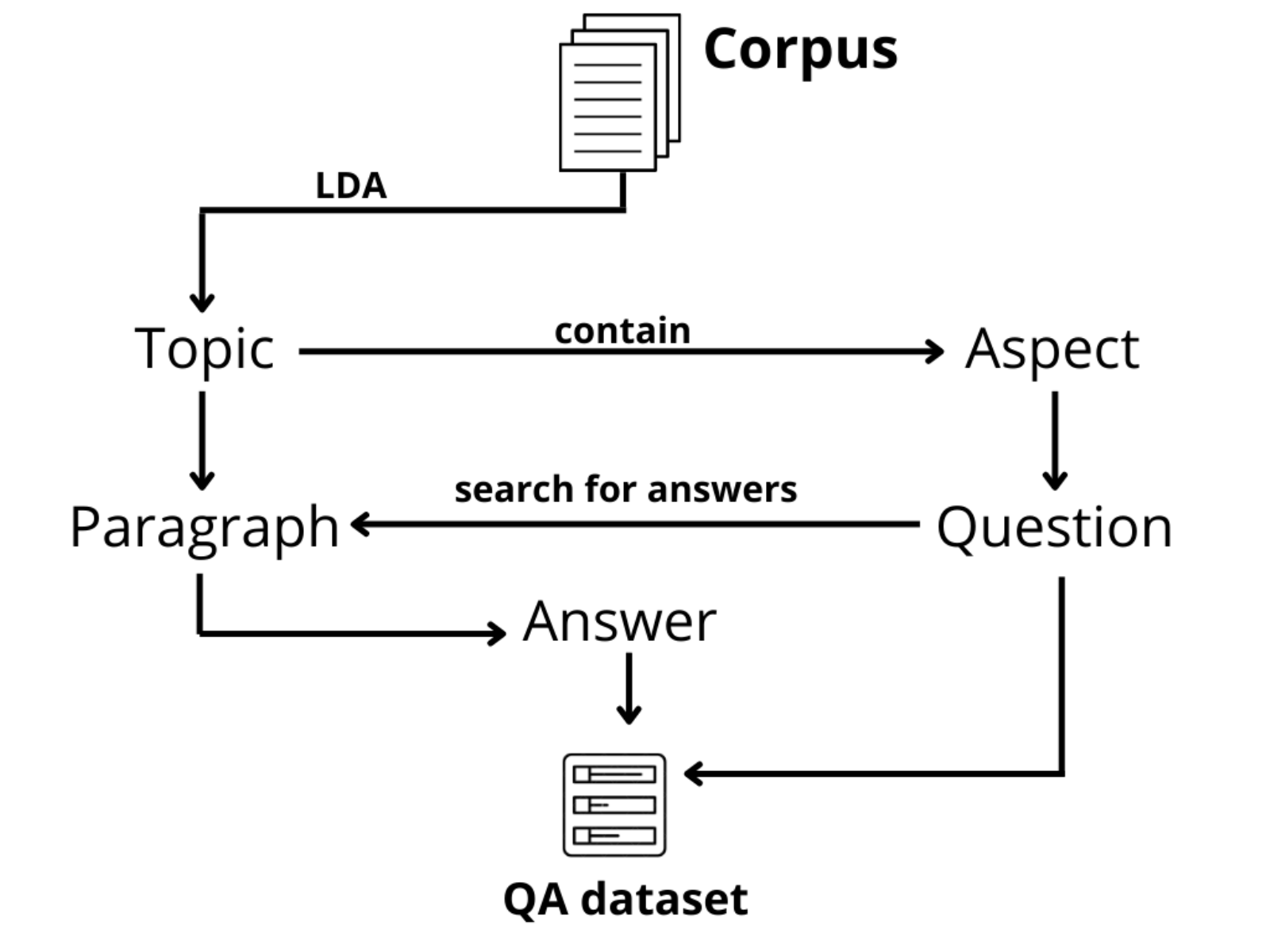} 
    \caption{Diagram to create a QA Dataset with LDA.}
    \label{fig:my_label}
\end{figure*}

In this way, it was possible to base the annotation of questions and answers in the QA dataset on the results of topics analysis, speeding up and making the annotation process faster. It was also possible to annotate multiple meaningful answers for the same question, greatly improving the quality of the QA dataset.

\begin{table}
 \centering
\caption{Statistics of the schizophrenia QA dataset.}\label{tab1}
\begin{tabular}{|l|l|}
\hline
Element & Count \\
\hline
    User posts & 415602  \\
    Max seq words & 385\\
    Topics paragraph & 35 \\
    Types of questions & 35\\
    Question and Answer & 1050\\
\hline
\end{tabular}
\end{table}

Table 2 shows the statistics of the proposed dataset. It contains 415602 posts grouped into 35 paragraphs (topics clusters), a maximum word sequence of 385 and a total of 1050 annotated questions and answers.

\subsection{QA pipeline}
To provide adequate input to the QA model, the corpus has been divided into paragraphs obtained through topic analysis processed. The document was converted to an appropriate input for the model by performing various cleaning operations. Normalised 3 consecutive blank lines so that they are only two blank lines, removed any white space at the beginning or end of each line of text, removed any long header or footer text that repeats on each page, ensured that document boundaries do not fall in the middle of sentences, and set the amount of overlap between two adjacent documents after a split. 

These data pre-processing steps had a great impact on the performance of the model to ensure optimal performance. The length of the documents also has a direct impact on the speed of the QA, having been divided into paragraphs. If the length of documents is halved, the workload of the reader is reduced.
The two main components of the QA pipeline are the \textit{Reader} and the \textit{Retriever}:
\begin{itemize}
    \item The \textbf{Reader} is the component that performs the closest analysis of the text, with great attention to syntactic and semantic detail, in order to find the span that best satisfies the question or query\cite{c7}. 
    \item The \textbf{Retriever} is a lightweight filter that can quickly go through the entire document archive and pass a set of candidate documents that are relevant to the query. It can sift through irrelevant documents, saving the Reader from doing more work than necessary and speeding up the query process. In addition, having divided the corpus into 35 paragraphs related topics provides an easier way of filtering the files to the reader, which is the central component that enables the QA model to find the answers\cite{c7}.
\end{itemize}
The top-k parameter in both Retriever and Reader determines how many results they return. More specifically, Retriever top-k dictates how many retrieved documents are passed to the next stage, while Reader top-k determines how many response candidates to show.
The choice of Retriever top-k is a trade-off between speed and accuracy, setting it higher means passing more documents to the Reader, thus reducing the chance that the answer-containing passage is missed\cite{c7}. Passing more documents to the Reader will create a larger workload for the component. In the experiments, it was found that Retriever top k=35 as the number of paragraphs and Reader top k=10 provided best performance.

\subsection{Experimental Setup}\label{sec:setup}
The BERT models were chosen for their ability to be easily adapted to the specific domain of interest \cite{c6,c7,c6,c9}. The models DistilBERT, RoBERTa, BioBERT were fine-tuned and compared with their pre-trained versions. 

\textbf{DistilBERT} is a transformers model, smaller and faster than BERT, which was pretrained on the same corpus in a self-supervised fashion, using the BERT base model as a teacher\cite{c29}. This means it was pretrained on the raw texts only, with no humans labelling them in any way with an automatic process to generate inputs and labels from those texts using the BERT base model. DistilBERT pretrained on the same data as BERT, which is BookCorpus, a dataset consisting of 11,038 unpublished books and English Wikipedia\cite{c29}.

\textbf{RoBERTa} is a transformers model pretrained on a large corpus of English data in a self-supervised way. This means it was pretrained on the raw texts only, with no humans labeling them in any way, with an automatic process to generate inputs and labels from those texts\cite{c24}. The RoBERTa model was pretrained on the reunion of five datasets: BookCorpus a dataset consisting of 11,038 unpublished books; English Wikipedia excluding lists, tables, and headers; CC-News a dataset containing 63 million English news articles crawled between September 2016 and February 2019; OpenWebText, an opensource recreation of the WebText dataset; Stories, a dataset containing a subset of CommonCraw data filtered to match the story-like style of Winograd schemas. Together, these datasets weight 160GB of text.

\textbf{BioBERT} (Bidirectional Encoder Representations from Transformers for Biomedical Text Mining), is a domain-specific language representation model pre-trained on large biomedical corpora.
BioBERT vastly outperforms BERT and previous state-of-the-art models in a variety of biomedical text mining tasks when pre-trained on biomedical corpora.
In our experiments, this model is used as benchmark for our comparison, since it is a model trained on medical data that achieves results in medical tasks equal to the current state of the art \cite{c39}. The model was trained on all MIMIC notes for 150,000 steps \cite{c39}.

To perform domain adaptation, the models are trained on the Schizophrenia QA dataset. The models were trained on 900 questions and answers annotated on the specific domain QA dataset for schizophrenia. The questions represented the entirety of the 35 topics previously identified, Utilizing 70\% of the QA dataset.
Table 3 shows the settings and parameters used for training the model.

\begin{table}
 \centering
\caption{Training parameters.}\label{tab1}
\begin{tabular}{|l|l|}
\hline
Parameter &  Value \\
\hline
        Optimizer   &  ADAM\\
        Weight decay    &  0.01  \\
        Learning rate   &  1e-05     \\
        Num of train epochs & 20 \\
        Max seq length  & 384 \\
        Max length  after clipping  & 256 \\
        Average length after clipping   &   255.90 \\
        Examples in train & 386001  \\ 
\hline
\end{tabular}
\end{table}

\subsection{Evaluation metrics}
The performance of the models used was judged through the Precision (1), Recall (2), F1 score (3), and Exact Match (EM).

The \textbf{Precision}, indicates the confidence of the model for each answer given; the values are in the range of [0 ; 1]. The intuition behind this score is as follows: a model that has on average a confidence score of 0.9 will provide correct predictions in about 9 out of 10 cases \cite{c7}.

The \textbf{F1score} is one of the best-suited metrics for this analysis, as it considers not only the number of prediction errors, but also the type of errors that are made. 
The F1 score is defined as the harmonic mean of precision and recall:
\begin{equation}
    \label{eq:eg2}
    Precision =  \frac{TP}{TP+FP} 
\end{equation}
\begin{equation}
    \label{eq:eg2}
     Recall =  \frac{TP}{TP+FN} 
\end{equation}
\begin{equation}
    \label{eq:eg2}
    F1 =  2*\frac{Precision*Recall}{Precision+Recall} 
\end{equation}
\\
Where TP is the number of True Positive, FP is the number of False Positive and FN is the number of False Negative.
The value of the F1 score is bounded in  $[0,1]$; if it is closer to 0 it means that precision and recall are low, if it is closer to 1 it means that precision and recall are high. 

The \textbf{EM-Exact Match} measures the proportion of cases in which the predicted answer is identical to the correct answer. For example, for the question-answer pair  “What is a schizophrenic afraid of” +  “He is afraid to leave the house”, a predicted answer such as ”He is afraid of leaving the house” produces a score not equal to 1 because there is no 100\% match.

The \textbf{Recall} was used to evaluate the Retriever in order to know whether the document containing the right answer is among the candidates. It is measured how many times the correct document was among the retrieved documents. For a single query, the output is binary: either a document is contained in the selection, or it is not. On the entire dataset, the recall score is a number between zero (no query retrieved the right document) and one (all queries retrieved the right documents).

\subsection{Result and evaluation}\label{sec:result}

Several experiments were carried out fine-tuning 3 different pre-trained models: DistilBERT, RoBERTa and BioBERT. 
The results are shown in Table 4.

\begin{table}
 \centering
\caption{Experiments results.}\label{tab1}
\begin{tabular}{|l|l|l|l|l|}
\hline
\textit{Models} & \textit{Precision} & \textit{F1} & \textit{Recall} & \textit{EM}\\
\hline
        DistilBERT &  0.681 & 0.639 & 0.713 & 0.425 \\
        RoBERTa &  0.746 & 0.743 & 0.7724 & 0.425 \\
        BioBERT &  0.790 & 0.775 & 0.880 & 0.427 \\
        Fine-tuned-DistilBERT  &  0.805 & 0.783 & 0.831 & 0.438  \\
        Fine-tuned-RoBERTa &  0.885 & 0.803 & 0.861 & 0.508\\
        Fine-tuned-BioBERT &  \textbf{0.903} & \textbf{0.885} &  \textbf{0.916} & \textbf{0.617}  \\
\hline
\end{tabular}
\end{table}

The fine-tuned-BioBERT model with the schizophrenia dataset achieved the best performance in our experiments.
There is an 14,304\% increase in the Precision score considering the best answer for each question between the BioBERT model and its fine-tuned version, 14,194\% for the F1 score, 10,229\% for the Recall score and 82,201\% for the EM score.
There is a boost of 38,498\% in terms of Precision score between the DistilBert model and the fine-tuned BioBERT model.
Moreover, for the Retriever there is a boost of 28,471\% in terms of Recall, between the DistilBert model and the fine-tuned model with BioBERT model. Using the pre-processed corpus divided into paragraphs with the LDA.
It is clear to notice how the fine-tuning of the models, with the presented schizophrenia dataset, improves the performance, compared to theirs pre-trained counterparts.

Examples of interaction with the fine-tuned BioBERT model are shown in Table 5. The questions asked concern the patient's daily problems and symptoms. It can be seen how specific data regarding patients with schizophrenia participating in the forum can be obtained. By judging the responses provided by the model, it is possible to conclude that the reported responses are relevant and make logical sense and as observable the model can provide not just one, but multiple relevant answers for each individual question.

\begin{table}
 \centering
\caption{Question-Answer examples obtained with the fine-tuned-BioBERT model.}\label{tab1}
\begin{tabular}{|l|l|}
\hline
   Question & Answer \\
\hline
  What is a schizophrenic afraid of?    & sanity/outside/ hallucinations. \\ 
  What is a schizophrenic obsessed with?    & food/conspiracy theories/rabies.\\ 
  What is a schizophrenic suffering from? & being tired/insomnia/delusions.\\
  What does a schizophrenic stop with? & smoking/ vomiting/drinking.  \\ 
  What does a schizophrenic struggle with? & hallucinations/medication/people. \\
  What does a schizophrenic see in hallucinations? & Demons/ spiders/ Pokémon.  \\ 
  
\hline
\end{tabular}
\end{table}

\section{Conclusion and Future works}\label{sec:conclusion}
The use of a noise-free data source specific not only to the medical macro domain, but also to the individual disease (in this case schizophrenia), it is of crucial relevance for the QA model. The use of web scraping applied to forums can be a new and valuable way to conduct data mining in the mental disorders field.

We demonstrated how through the Latent Dirichlet Allocation (LDA) method is possible to identify the main topics and aspects that can be used for defining questions and answers in the QA dataset. It is possible to obtain more meaningful and multiple answers for each individual question and successfully accelerating the question-answer annotation process.

We created a QA dataset dedicated to schizophrenia with 415602 user posts, 35 paragraphs, 35 different type questions and 1050 answers.
We showed how to preprocess the corpus and how \emph{Reader} and \emph{Retriever} can be best used for the extraction of information regarding the experiences and symptoms of forum users suffering from schizophrenia. In our experiments, using and adapting BERT language models, we achieved results equal to the current state of the art in the domain of QA models for mental disorders.

A possible future goal will be to build a sufficiently large domain-specific QA dataset for mental disorders. Thousands of annotations could make a difference in terms of the quality of the responses obtained, especially from the perspective of reusing such data for different mental disorders.
Another future work could be the application of a Question Generator, which takes as input a document and generates questions that it believes can be answered in the \cite{c41} document. It is almost the reverse of the reader, which takes a question and a document as input and returns an answer.
In addition, it is possible to think of applying the methodology shown in this paper to different areas. In fact, the number of currently active forums is huge and very various.

%
%
%
%

\end{document}